\documentclass[sigconf]{acmart}

\usepackage{subfigure}
\usepackage{algpseudocode}
\usepackage{algorithmicx,algorithm}
\usepackage{multirow}

\usepackage{balance}

\AtBeginDocument{%
  }

\copyrightyear{2024}
\acmYear{2024}
\setcopyright{rightsretained}
\acmConference[WWW '24]{Proceedings of the ACM Web Conference 2024}{May 13--17, 2024}{Singapore, Singapore} \acmBooktitle{Proceedings of the ACM Web Conference 2024 (WWW '24), May 13--17, 2024, Singapore, Singapore}\acmDOI{10.1145/3589334.3645602}
\acmISBN{979-8-4007-0171-9/24/05}

\makeatletter
\gdef\@copyrightpermission{
  \begin{minipage}{0.3\columnwidth}
   \href{https://creativecommons.org/licenses/by/4.0/}{\includegraphics[width=0.90\textwidth]{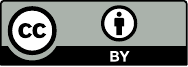}}
  \end{minipage}\hfill
  \begin{minipage}{0.7\columnwidth}
   \href{https://creativecommons.org/licenses/by/4.0/}{This work is licensed under a Creative Commons Attribution International 4.0 License.}
  \end{minipage}
  \vspace{5pt}
}
\makeatother

\begin{document}

\title{RulePrompt: Weakly Supervised Text Classification with Prompting PLMs and Self-Iterative Logical Rules}

% \author{}
% \authornote{Corresponding Author}
% % \authornote{123}
% \affiliation{%
%   \institution{$^{1}$Institute of Software, Chinese Academy of Sciences, Beijing, China \\}
%   \institution{$^{2}$University of Chinese Academy of Sciences, Beijing, China \\}
%   \country{}
%   School of Computing and Artificial Intelligence
  % \streetaddress{P.O. Box 1212}
  % \city{Yangzhou}
  % \state{Jiangsu}
  % \country{China}


% % \author{Miaoimiao Li\inst{1,2,3}}
% % \authornote{Both authors contributed equally to this research.}
% % \email{trovato@corporation.com}
% % \orcid{1234-5678-9012}
% \author{Miaomiao Li\inst{1}}
% \email{limiaomiao22@mails.ucas.ac.cn}
% \affiliation{%
%   \institution{Institute for Clarity in textation}
%   % \streetaddress{P.O. Box 1212}
%   % \city{Dublin}
%   % \state{Ohio}
%   % \country{USA}
%   % \postcode{43017-6221}
% }

\author{Miaomiao Li}
\affiliation{%
  \institution{Institute of Software, Chinese Academy of Sciences, Beijing, China}
  \institution{University of Chinese Academy of Sciences, Beijing, China}
  \country{}
  }
\email{limiaomiao22@mails.ucas.ac.cn}

\author{Jiaqi Zhu}
\authornote{Corresponding Author}
\affiliation{%
  \institution{Institute of Software, Chinese Academy of Sciences, Beijing, China}
  \institution{University of Chinese Academy of Sciences, Beijing, China}
  \country{}
  }
\email{zhujq@ios.ac.cn}

\author{Yang Wang}
\affiliation{%
  \institution{Institute of Software, Chinese Academy of Sciences, Beijing, China}
  \institution{University of Chinese Academy of Sciences, Beijing, China}
  \country{}
  }
\email{wangyang223@mails.ucas.ac.cn}

\author{Yi Yang}
\affiliation{%
  \institution{Institute of Software, Chinese Academy of Sciences, Beijing, China}
  \country{}
  }
  
\author{Yilin Li}
\affiliation{%
  \institution{Institute of Software, Chinese Academy of Sciences, Beijing, China}
  \institution{University of Chinese Academy of Sciences, Beijing, China}
  \country{}
  }

\author{Hongan Wang}
\affiliation{%
  \institution{Institute of Software, Chinese Academy of Sciences, Beijing, China}
  \institution{University of Chinese Academy of Sciences, Beijing, China}
  \country{}
  }

% \author{Miaomiao Li}
% \affiliation{%

% \affiliation{%

% \affiliation{%

% \affiliation{%

% \affiliation{%

% \affiliation{%

% \affiliation{%

\renewcommand{\shortauthors}{Miaomiao Li et al.}


\begin{abstract}

Weakly supervised text classification (WSTC), also called zero-shot or dataless text classification, has attracted increasing attention due to its applicability in classifying a mass of texts within the dynamic and open Web environment, since it requires only a limited set of seed words (label names) for each category instead of labeled data. With the help of recently popular prompting Pre-trained Language Models (PLMs), many studies leveraged manually crafted and/or automatically identified verbalizers to estimate the likelihood of categories, but they failed to differentiate the effects of these category-indicative words, let alone capture their correlations and realize adaptive adjustments according to the unlabeled corpus. In this paper, in order to let the PLM effectively understand each category, we at first propose a novel form of rule-based knowledge using logical expressions to characterize the meanings of categories. Then, we develop a prompting PLM-based approach named RulePrompt for the WSTC task, consisting of a rule mining module and a rule-enhanced pseudo label generation module, plus a self-supervised fine-tuning module to make the PLM align with this task. Within this framework, the inaccurate pseudo labels assigned to texts and the imprecise logical rules associated with categories mutually enhance each other in an alternative manner. That establishes a self-iterative closed loop of knowledge (rule) acquisition and utilization, with seed words serving as the starting point. Extensive experiments validate the effectiveness and robustness of our approach, which markedly outperforms state-of-the-art weakly supervised methods. What is more, our approach yields interpretable category rules, proving its advantage in disambiguating easily-confused categories.

\end{abstract}




%% information and builds the first part of the formatted text.
\maketitle

\section{Introduction}

With the rapid development of Internet, an abundance of textual content is produced across news media and social networks. It is significant and challenging to classify these texts into predefined categories, especially when up-to-date labeled data are hard to access due to the dynamic and open nature of the Web. Consequently, there has been a growing interest in weakly supervised text classification (WSTC) \cite{zhang2022motifclass, wang2021x, li2023CL-WSTC, zhang2023PIEClass, wang2023WSTCBenchmark, meng2018weakly}, also known as zero-shot or dataless text classification \cite{chang2008importance,  song2014dataless, chen2015dataless, li2016effective, li2019filtering, yang2021dataless, yang2021effective, meng2018weakly, mekala2020contextualized, meng2020text, zhang2021weakly, wang2023PESCO, zhao2023NPPrompt}, which only requires a limited set of seed words (label names) for each category.

Recently, the proliferation of prompting Pre-trained Language Models (PLMs) greatly bolstered the WSTC task, but their performances still lag behind supervised methods \cite{wang2023WSTCBenchmark}. Since no labeled data are available as evidence, relying solely on seed words for grasping category meanings proves inadequate. In previous researches, many approaches either provided manual verbalizers of categories or automatically discovered them based on word embedding similarity. Taking them as additional knowledge, some studies estimated category likelihoods by tapping into the generative capability of PLMs \cite{zhao2023NPPrompt, zhang2023PIEClass}, and others leveraged PLM's effective vector representations to calculate the similarity or entailment between texts and categories \cite{wang2023PESCO, park2022lime}. However, most of them failed to differentiate the effects of these category-indicative words (abbreviated as indicative words). Although NPPrompt \cite{zhao2023NPPrompt} did calculate and utilize the weights of them, their roles in classification remained independent of each other and lacked adaptive adjustments for the current corpus, so cannot accommodate ever-changing Web environment.

However actually, the effect of each category-indicative word varies and is worth further explorations. Certain words can determine the category on its own, like the label names, while others need to be used cooperatively to distinguish between easily-confused categories.
For example, the word ``\textit{penalty}'' itself cannot signify the ``\textit{Sports}'' category, but when combined with ``\textit{goal}'', the text is likely to talk about a football match. Conversely, an additional word ``\textit{company}'' could imply the ``\textit{Business}'' category rather than ``\textit{Sports}''. Therefore, a simplistic set of indicative words is not enough to cover the full meanings of categories. Instead, logical operations such as conjunction and disjunction are appropriate to capture the correlations of these words as enriched and precipitable knowledge for weakly supervised classification. Luckily, the flexibility of prompting PLMs just offers an opportunity to apply these logical rules in the template to achieve precise semantic representations of categories.

It is obvious that logical rules are difficult to set manually as prior knowledge, but they can be mined from preliminarily categorized texts with the aid of pseudo labels generated by the PLM. Furthermore, the mined rules and pseudo labels can mutually enhance each other in an alternative way, establishing a self-iterative closed loop for knowledge acquisition and utilization, with seed words as the starting point. That poses two main challenges: (1) When inaccurate pseudo labels are available, how to identify candidate category-indicative words using the PLM and build correlations among them by means of logical rules to characterize each category? (2) With imprecise logical rules, how to effectively transform them into the PLM template for classification by handling each logical operator discriminatively, and then update the pseudo label assigned to each text? 

To address these issues, this paper at first proposes a novel kind of rule-based knowledge in the form of logical expressions for category understanding in WSTC. Each category is represented by a disjunctive normal form, where indicative words serve as atomic propositions. Specifically, a single disjunctive term (one-literal clause) denotes strong and self-explanatory indicative words, while a clause of conjunctive form depicts the synergistic effect of weak and polysemous indicative words.

Based on this, a prompting PLM-based approach for text classification is developed, through iteratively updating both the pseudo label of each text and the logical rule of each category. That is realized mainly via two modules, rule mining and rule-enhanced pseudo label generation. The former first extracts signal words from each text by the PLM, and then regards these words as a transaction of the relevant category decided by the current pseudo labels. For each category, we mine frequent 1-itemsets (items) and 2-itemsets respectively from specific subsets of transactions, and construct the disjunctive normal form. In the latter module, the current logical rule for each category is injected into three PLM-based models, each providing a different perspective. Afterwards, a new pseudo label is generated for each text via integrating the results of these models. In addition, in each iteration, the PLM can be fine-tuned with a self-supervised loss to better align with the task requirements.

In summary, the contributions of this paper include:
\begin{itemize}
    \item To the best of our knowledge, this is the first attempt to differentiate the effects of category-indicative words in the WSTC task and characterize category meanings through logical rules, thereby establishing a new paradigm for knowledge representation in this field.
    \item A novel approach leveraging prompting PLMs is presented to make the pseudo labels of texts and the logical rules of categories enhance each other iteratively. That facilitates a sufficient fusion of automatically generated rule-based knowledge and unlabeled data.
    \item Comprehensive experiments conducted on multiple real datasets demonstrate the effectiveness and interpretability of our approach. It consistently outperforms state-of-the-art weakly supervised methods, and yields intuitive logical rules for categories to avoid confusion.
\end{itemize}

\section{Related Work}

\subsection{Weakly Supervised Text Classification}

Weakly supervised text classification (WSTC) demands minimal seed information, such as label names or extended keywords for each category, thereby significantly reducing the cost of text annotations.
At an early stage, some researchers used auxiliary knowledge bases like Wikipedia to establish the semantic correlation between texts and labels \cite{chang2008importance, song2014dataless}.
Subsequently, topic-model based methods emerged \cite{chen2015dataless, li2016effective, li2019filtering, yang2021dataless, yang2021effective}, which inferred category-aware topics from a limited set of seed words. 
In the last few years, neural methods has gained prominance \cite{meng2018weakly, mekala2020contextualized, zhang2021weakly, wang2021x, zhang2022motifclass}. They trained neural classifiers using pseudo labels of texts, often relying on generated pseudo-texts or PLMs to detect category-indicative keywords. For example, LOTClass \cite{meng2020text} used label names as the only seed words, and introduced BERT for category understanding. 

In recent time, prompt-based methods \cite{hu2022knowledgeable, fei2022beyond, park2023cross} have been extensively developed for the WSTC task. 
A lot of work harnessed the strong generative capability of PLMs with instruction templates for classification.
For instance, NPPrompt \cite{zhao2023NPPrompt} used initial word embeddings by PLM to automatically construct verbalizers without manual design or unlabeled corpus, and estimated the probability distribution over categories through weighted sum of these words.
PIEClass \cite{zhang2023PIEClass} introduced a noise-robust method to iteratively self-train text classifiers and update pseudo labels, employing two fine-tuning strategies of PLMs to improve the quality of pseudo labels. 
WDDC \cite{zeng2022weakly} utilized the generated words at the [MASK] token as supervision signals, and proposed a latent variable model to train a word distribution learner and a text classifier simultaneously.
Other approaches explored the vector representation power of prompting PLMs.
PESCO \cite{wang2023PESCO} incorporated label descriptions into predefined prompts, formulating the WSTC task as a neural matching problem. 
Meanwhile, LIME \cite{park2022lime} used large textual entailment models trained with external data to suggest seed words and infer text labels.

Although these methods have demonstrated inspiring performances, a gap still exists when compared to fully supervised methods.
Due to the absence of labeled data, there is a notable need to automatically extract and apply additional knowledge from unlabeled data during the classification process. Existing methods just relied on a set of category-indicative words, but have not taken the varying effect of these words into account, which leads to imprecise category understanding.

\subsection{Logical Rules for Natural Language Processing Tasks}

Recently, there have been increasing researches on the integration of logical rules into natural language processing tasks, aiming to improve the interpretability of neural network models.

Hu et al. \cite{hu-etal-2016-harnessing} proposed a teacher-student framework combining deep neural networks with first-order logic rules, and transformed the structured information of logic rules into the weights of neural networks.
TALLOR \cite{li-etal-2021-weakly} addressed the named entity tagging problem by using a small set of seed logical rules as weak supervision, and further selected new accurate logical rules based on a hand-tuned threshold.
PTR \cite{han2022ptr} incorporated logic rules to encode human prior knowledge and composed several manually designed sub-prompts into final task-specific prompts.
PRBoost \cite{zhang-etal-2022-prompt} viewed the top-$k$ predictions at the [MASK] token of large-error instances as candidate rules through the disjunction operation, and then used human-selected ones to generate weak labels for model training.

However, most of these previous work required seed rules as initial supervision or human feedback when selecting accurate rules. In contrast, our approach focuses on the WSTC task, and establishes self-iterative closed loop for the acquisition and utilization of logical rules, eliminating the need for human intervention. Additionally, while existing PLM-based methods primarily employed single operator when composing decision rules, we leverage both the disjunction and conjunction operators to distinguish the strength and effect of indicative words, enabling a more precise understanding of categories.

\section{Preliminaries}

In this section, we formulate the task of weakly supervised text classification (WSTC), and briefly introduce prompting PLMs as well as two roles of them as the foundation of our approach.

\subsection{Problem Formulation}

Given a corpus of unlabeled texts $D = \{D_1,\ldots,D_N \}$ and a set of target categories $Z=\{z_1, \ldots, z_K\}$ with a label name $l(z)$ for each $z\in Z$,  weakly supervised text classification (WSTC) aims to assign a category label $z(d)$ to each text $d$.
Following the extremely weak supervision setting \cite{wang2021x}, only the sole label surface name of each class is used as supervision here, without other seed words.

\subsection{Prompting PLMs for Estimating Likelihoods}
\label{sec:pre2}

Prompt-based tuning applies cloze-style tasks to tune PLMs. 
A prompt is composed of a template $\mathcal{T}(\cdot)$ and a set $\mathcal{V}$ of selected words. 
We can fill each text $d$ into the template $\mathcal{T}(\cdot)$ to obtain the prompt input $\mathcal{T}(d)$.
For example, for the text classification task on news, the prompt can be written as:
\begin{equation}
\label{eq:MASK}
\mathcal{T}(d)=d {\rm~It~is~about~[MASK]~news.}
\end{equation}

In vanilla prompt engineering, the verbalizer, i.e.,
an injective mapping function $\phi: Z \rightarrow \mathcal{V}$, links the category set and the set of selected words. 
Then, at the masked position, we can calculate the likelihood for each category via word probability distributions:
\begin{eqnarray}
\label{eq:verba}
P(z|d)=P({\rm [MASK]}=\phi(z)~|~\mathcal{T}(d)).
\end{eqnarray}

Recently, A lot of work studied for a verbalizer with richer label words to represent the category.
Typically, NPPropmt \cite{zhao2023NPPrompt} constructs a $K$-nearest-neighbor verbalizer, through searching over the whole vocabulary $\mathcal{V}$ for the top-$k$ nearest words to the label name of $z$ in the embedding space of the PLM, denoted as $\mathcal{M}(z)$: 
\begin{eqnarray}
\label{eq:complement}
\mathcal{M}(z)= \mathop{{\rm Top-}K_0}_{v \in \mathcal{V}}\{{\rm sim}(\bold{emb}(v),\bold{emb}(l(z))\},
\end{eqnarray}
where $\bold{emb}(v)$ and $\bold{emb}(l(z))$ are the embedding vectors of word $v$ and label name $l(z)$ respectively, and ${\rm sim}(\cdot)$ means cosine similarity.

Then, we get the unnormalized probability for each category:
\begin{eqnarray}
\label{eq:Q}
Q(z|d)=  \displaystyle \sum_{v \in \mathcal{M}(z)} w(v,l(z)) \cdot \Theta({\rm [MASK]}=v~|~\mathcal{T}(d)),
\end{eqnarray}
where $\Theta$ is the logit vector instead of probability for kernel smoothing, and $w(v,l(z))$ is the weight of the word $v$ on the label name $l(z)$, defined in the softmax form:
\begin{eqnarray}
w(v,l(z))=\frac{{\rm exp}({\rm sim}(\bold{emb}(v),\bold{emb}(l(z))\})}{\sum_{v' \in \mathcal{M}(z)}{\rm exp}({\rm sim}(\bold{emb}(v'),\bold{emb}(l(z)))}.
\end{eqnarray}

Besides, NPPrompt uses more than one keywords for certain categories. The final score is calculated as follows:
\begin{eqnarray}
\label{eq:maxQ}
Q(z|d)= \mathop{{\rm max}}_{v\in \Phi(z)}Q(v|d)),
\end{eqnarray}
where $\Phi(z)$ contains all keywords for category $z$, and $Q(v|d)$ is computed similar to Equation~\ref{eq:Q}, replacing the category $z$ by one of its indicative words $v$ and the label name $l(z)$ just by $v$ itself.

\subsection{Prompting PLMs for Getting Signal Words}

In addition to estimating category likelihoods, some work \cite{zeng2022weakly} utilized prompting PLMs to generate words which can summarize the content of the given text.  
That also depends on the probability distribution over $\mathcal{V}$, and can be used to get better supervision information than the words themselves appearing in the text.
% which represent the probability that the corresponding words 
Formally, given a threshold $K_1$, for each text $d$, the top $K_1$ words with higher logits can be seen as signal words of $d$, denoted as $SW(d)$:
\begin{eqnarray}
\label{eq:SW}
SW(d)= \mathop{{\rm Top-}K_1}_{v \in \mathcal{V}}\{P({\rm [MASK]}=v~|~\mathcal{T}(d))\}.
\end{eqnarray}

\section{Method}

\begin{figure*}[ht]
  \centering
  \includegraphics[width=\linewidth,height=0.36\textwidth]{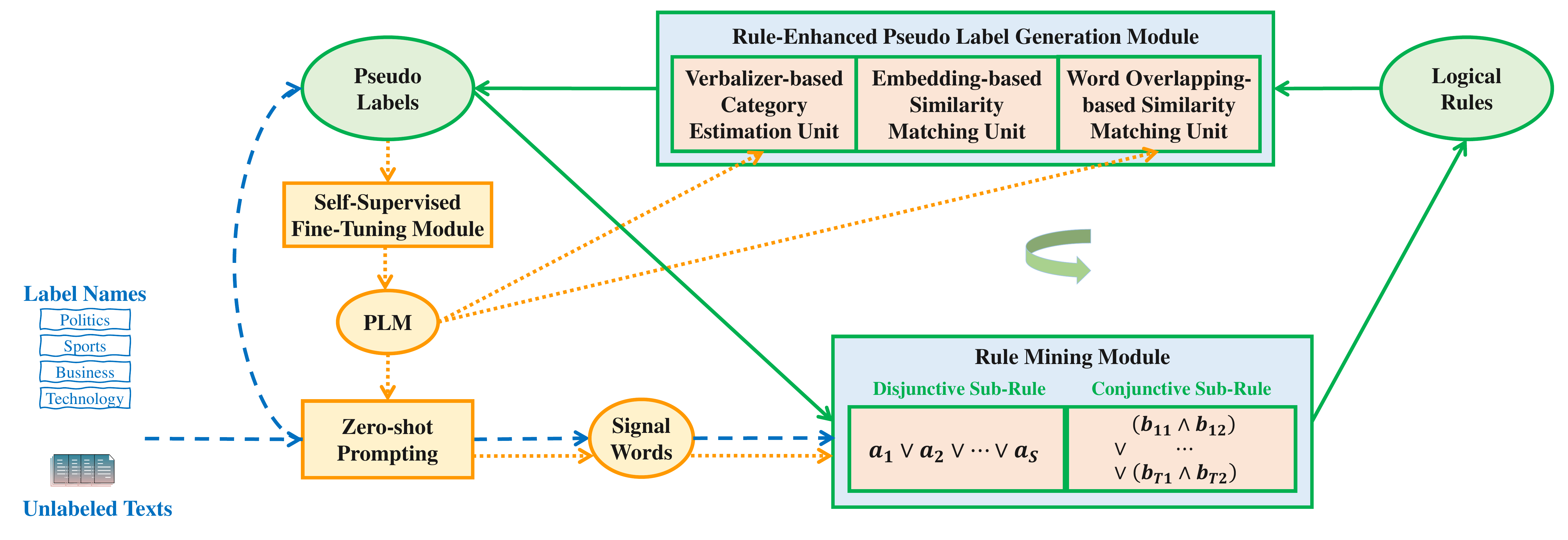}
  \caption{Framework of the Proposed Approach RulePrompt.}
  % \Description{XXX}
  \label{fig:framework}
\end{figure*}

In this section, we at first define logical rules of categories as a new kind of knowledge. Based on this, the framework of RulePrompt is presented followed by details of the three key modules.

\subsection{Logical Rules of Categories}

In this paper, we propose a novel kind of rule-based knowledge representation for categories, as additional weak supervision information in text classification. It takes automatically mined category-indicative words as atomic propositions, and build their correlations through logical expressions with disjunction and conjunction operators. Specifically, each category can be represented by a disjunctive normal form.

\begin{definition}[Logical Rules of Categories]\label{def:LR}
The meaning of each category $z$ can be represented by a logical rule as follows:
\begin{eqnarray}
r(z) = \Big( a_1 \vee \cdots \vee a_S \Big) \vee \Big( (b_{11} \wedge b_{12}) \vee \cdots \vee (b_{T1} \wedge b_{T2}) \Big), 
\end{eqnarray}
where both $a_j~(1\le j\le S)$ and $b_{j1},b_{j2}~(1\le j\le T)$ are indicative words of the category $z$. 
The rule can be divided into two sub-rules. The first $S$ words are strong and can indicate the category on its own, so they are connected directly by the disjunction operator and compose the disjunctive sub-rule, denoted as $r^{\rm d}(z)$. On the contrary, the last $2T$ words are comparatively weak and need to act together to imply the category, so they are firstly paired with the conjunction operator, and then combined by disjunction. That is called the conjunctive sub-rule and denoted as $r^{\rm c}(z)$. 

Despite the relations of indicative words above are not the same as those in classical logical rules, the ideas of conjunction and disjunction are actually utilized here to obtain precise semantic representations of categories in two views.
Notice that a simplified version of logical operations is adopted by restricting the conjunction on just two words. That is reasonable and empirically effective, since the discrepancy between individual words and two-word pairs is essential, compared to $n$-word sets ($n>2$).
\end{definition}

\subsection{Framework}

On this basis, we propose a novel prompting PLM-based approach for the WSTC task as shown in Figure~\ref{fig:framework}.
At first, as the starting point with only label names, we leverage a classical zero-shot prompting method using PLM \cite{zhao2023NPPrompt} to generate the initial pseudo labels and the signal words of texts (blue dashed line).
Then, the approach enters the self-iteration between pseudo labels and category knowledge (logical rules) through mutual enhancement (green solid line). Meanwhile, the PLM is gradually optimized by self-supervised fine-tuning to adaptively support the main iteration above (yellow dotted line). 
To achieve the whole process, three modules are designed.

In the rule mining module, based on the current pseudo labels with confidence scores, we cluster the unlabeled texts assigned to each category into three sets. Then, with the signal words of each text obtained by PLM, frequent 1-itemsets (items) and 2-itemsets of each category are mined from the first two confident sets respectively, which composes the disjunctive normal form of the logical rule for each category.

In the rule-enhanced pseudo label generation module, we incorporate the current logical rules into three prompting PLM-based classification models from different perspectives to update pseudo labels. 
On the one hand, the words in the disjunctive sub-rule with higher support is directly used to obtain a richer verbalizer in a generation-based model.
On the other hand, the whole rule is injected into templates to derive texts for similarity-based classification.
That is realized in two views, global embedding similarity and local word overlapping. Finally, these results are averaged to get new pseudo labels of texts. 
s
Moreover, in order to make the PLM accommodate this specific task, the self-supervised fine-tuning module is executed after each time pseudo labels are generated, employing self-supervised loss over high-confidence texts.

\subsection{Rule Mining Module}

In the weakly supervised setting, only label names are not adequate to reflect the meanings of categories. Thanks to the strong generative and representation capability of prompting PLMs, it is feasible to utilize the pseudo labels and signal words of texts to furthermore understand categories and enrich the prior knowledge.
Since pseudo labels are imperfect, for the sake of mitigating error propagation, the selection of texts and signal words should be restricted to those with high confidence. Inspired by previous work \cite{liao2022zero,budd2021survey,scheffer2001active}, 
we define the confidence score (for the pseudo label) of a text as: 
\begin{eqnarray}
\label{eq:conf}
conf(d)=P(z_{(1)}|d)-P(z_{(2)}|d),
\end{eqnarray}
where $z_{(1)}$ and $z_{(2)}$ respectively denote the first and the second most probable label for text $d$ computed by the prompting PLM. 
Compared to the highest probability, this difference value gives a better indication of how confident the PLM regards the current unique prediction. 

However, for each category $z$, the numbers of texts appropriate to extract strong words and weak words are hard to determine, so we adaptively cluster the texts assigned to $z$ into three sets via K-means, based on the confidence scores. These texts with excellent, good and poor quality, are denoted as $D^1_z$, $D^2_z$ and $D^3_z$ respectively.

For the signal words of texts, the set $SW(d)$ computed by Equation~\ref{eq:SW} needs to be further filtered to guarantee their competence as indicative words. To this end, we utilize the whole corpus to pursue the speciality of signal words for the text, which we think can better imply the assigned category as well.
The new unnormalized probability can be calculated as:
\begin{eqnarray}
P'({\rm[MASK]}=v~|~\mathcal{T}(d))=\frac{P({\rm[MASK]}=v~|~\mathcal{T}(d))}
{\frac{1}{N} \displaystyle \sum_{d' \in D}P({\rm[MASK]}=v~|~\mathcal{T}(d'))}.
\end{eqnarray}
 
Then, we select the top $K_2$ signal words with higher logits as the strong signal words, denoted as $SSW(d)$:
\begin{eqnarray}
\label{eq:SSW}
SSW(d)= \mathop{{\rm Top-}K_2}_{v \in \mathcal{V}}\{P'({\rm [MASK]}=v~|~\mathcal{T}(d))\}.
\end{eqnarray}

\begin{figure}[t]
  \centering
  \includegraphics[width=\linewidth]{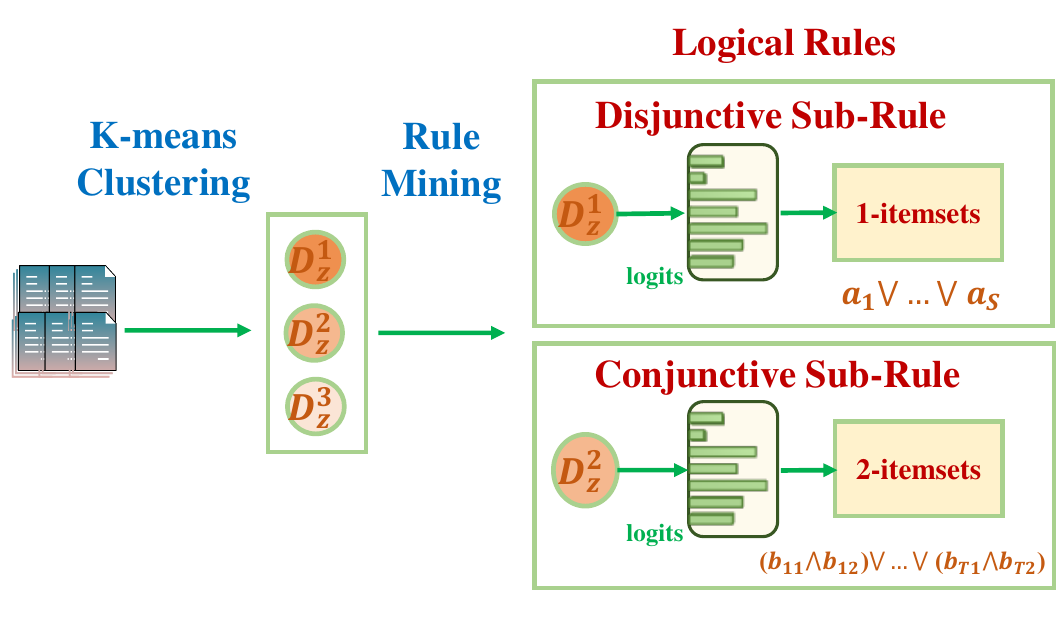} %rule
  \caption{Rule Mining Module.}
  % \Description{aaa.}
  \label{fig:rule}
\end{figure}

Next, we use frequent pattern mining \cite{agrawal1994fast, han2000FreeSpan, pei2004PrefixSpan} to obtain representative rules of categories.
For $D^1_z$ and $D^2_z$, we treat each text as a transaction and each strong signal word of it as an item of the transaction.
We at first pay attention to the most confident set $D^1_z$ to mine frequent 1-itemsets (items) with a predefined support threshold $h_1$, which compose the disjunctive sub-rule of $z$, as each of them alone is enough to indicate a category. The support of a word $a$ in $D^1_z$ is calculated as:
\begin{equation}
\label{eq:sup1}
sup(a,D^1_z)= \frac{\sum_{d\in D^1_z} I_1(a,d)}{|D^1_z|},
\end{equation}
where $I_1(a,d)$ is an indicator function expressing whether $a$ is in the transaction of $d$, i.e.,
\begin{equation}
I_1(a,d)= \left\{
\begin{aligned}
1, a \in SSW(d),  \\
0, a \not\in SSW(d).
\end{aligned}
\right.
\end{equation}

In addition, for the set $D^2_z$ with moderate confidence scores, we mine 2-itemsets given another threshold $h_2$ to construct the conjunctive sub-rule. Although these words cannot represent a category individually, their co-occurrence in the set of strong signal words should also be captured. 
The support of a 2-itemset $b=b_1\land b_2$ is calculated as:
\begin{equation}
\label{eq:sup2}
sup(b,D^2_z)= \frac{\sum_{d\in D^2_z} I_2(b,d)}{|D^2_z|},
\end{equation}
where $I_2(b,d)$ is another indicator function expressing whether both $b_1$ and $b_2$ are in the transaction of $d$, i.e.,
\begin{equation}
I_2(b,d)= \left\{
\begin{aligned}
1, b_1 \in SSW(d) \land b_2 \in SSW(d),  \\
0, b_1 \not\in SSW(d) \lor b_2 \not\in SSW(d).
\end{aligned}
\right.
\end{equation}

Besides, we need to exclude those pairs containing words also appearing in the frequent 1-itemsets of $D^2_{z'}$ for any other category $z'$, which would bring confusion.

\subsection{Rule-Enhanced Pseudo Label Generation Module}

In this subsection, we present the reversed direction of the iteration, i.e., how to inject the mined logical rules of categories into the pseudo labels generation process. Considering the diverse capabilities of PLMs and the distinct roles that logical rules play within them, three units from different perspectives are designed to compute the probability of each text belonging to each category.
Final results are obtained by averaging the outputs from the three units.

\subsubsection{Verbalizer-based Category Estimation Unit}

Since label names are too limited to characterize categories, the indicative words in our logical rules can be naturally used to expand the verbalizers in classical zero-shot prompting models (Equation~\ref{eq:verba}).
In view of the strictness of verbalizers, for each category, we only use the words in the first half of the disjunctive sub-rule according to their support values. The expanded set is written as $\Phi'(z)=\{l(z), a_1, a_2, \ldots, a_{\frac{S}{2}}\}$, which acts similarly with the manually crafted set of keywords in Equation~\ref{eq:maxQ}.
Besides, inspired by NPPrompt \cite{zhao2023NPPrompt}, the top-$K_0$ closest words to each of them are also used to complement the verbalizer (Equation \ref{eq:complement}). In this way, for a keyword $v\in \Phi'(z)$, we can get the probability $Q(v|d)$ and then take the maximum value among all keywords as the aggregated probability $Q(z|d)$ similar to Equation~\ref{eq:maxQ}, as all of these words can imply the category independently.

Noticing that $Q$ is an unnormalized probability, we use the softmax function to transform the value between 0 and 1, to get the normalized probability $P_1(z|d)$ from the first perspective:
\begin{equation}
\label{eq:P1}
P_1(z|d)= \frac{{\rm exp}(Q(z|d))}{\sum_{z' \in Z}{\rm exp}(Q(z'|d))}.
\end{equation}

\subsubsection{Embedding-based Similarity Matching Unit}

To conduct a similarity-based matching between a text and a category through prompting PLM, an intuitive idea is to put the logical rule of each category into the [MASK] token of the template in Equation~\ref{eq:MASK} to form a complete sentence \cite{wang2023PESCO,zhang-etal-2022-prompt}.
However, the expressions of conjunction and disjunction are not like natural language texts, which would affect the semantic understanding of the PLM.
Hence, we handle each indicative word separately instead and combine them in different ways for disjunction and conjunction.

For the disjunctive sub-rule, we directly calculate embedding-based similarity between a text $d$ and a category $z$ as weighted sum of the similarity between $d$ and each word $a$ in the sub-rule of $z$:
\begin{equation}
ES^{\rm d}(d,z) = \frac{\sum_{a\in r^{\rm d}(z)} sup(a,D^1_z) \cdot {\rm sim}(f(d),g(a))}{S},    
\end{equation}
% $$s(d,a)={\rm sim}(f(d),g(a)).$$
where $sup(a,D^1_z)$ is the support of word $a$ in $D^1_z$, $f(d)$ is the sentence embedding of text $d$, 
and $g(a)=f(\mathcal{T}'(a))$ is the embedding of the template after removing ``$d$'' and replacing [MASK] with $a$.

% $\mathcal{T}'(a)=$ It is about $a$ news.

While for the conjunctive sub-rule, besides that the outer disjunction operations can be handled in the same way, the similarity between $d$ and each 2-itemset $b=b_1\land b_2$ is computed through the weighted composition of vectors instead of similarity scores:
\begin{equation}
ES^{\rm c}(d,z) = \frac{\sum_{b\in r^{\rm c}(z)} sup(b,D^2_z) \cdot {\rm sim}(f(d),g'(b))}{T},    
\end{equation}
% $$s(d,b)={\rm sim}(f(d),g(b)).$$
where $sup(b,D^2_z)$ is the support value of the 2-itemset $b$ in $D^2_z$, and the embedding variant $g'(b)$ can be computed as the weighted sum of the embedding vectors of two conjunctive terms of $b$, utilizing the 1-itemset support of them in $D^2_z$:
\begin{eqnarray}
g'(b) & = & \frac{sup(b_1,D^2_z)}{sup(b_1,D^2_z)+sup(b_2,D^2_z)} \cdot f(\mathcal{T}'(b_1)) \nonumber \\
 & & +~~\frac{sup(b_2,D^2_z)}{sup(b_1,D^2_z)+sup(b_2,D^2_z)} \cdot f(\mathcal{T}'(b_2)).
\end{eqnarray}

At last, the embedding-based similarity $ES(d,z)$ between $d$ and $z$ is defined as the maximum value for the two sub-rules, and regarded as the probability $P_2(z|d)$ from the second perspective:
\begin{equation}
P_2(z|d)= ES(d,z) =  {\rm max} (ES^{\rm d}(d,z), ~ ES^{\rm c}(d,z)).
\end{equation}

\subsubsection{Word Overlapping-based Similarity Matching Unit}

Following the idea of PRBoost \cite{zhang-etal-2022-prompt}, besides embedding-based similarity matching in a global view, we also examine the word overlapping-based similarity in a local view, leveraging PLMs' capability of generating signal words from texts once again.
In this way, the rule is no longer inserted into the [MASK] token, but occupies the position of the input text in the template as an independent sentence.
Consequently, the word-level rule needs to be transformed to a coherent sentence with the help of text connectors. Considering the typical human speech patterns, we use the word ``and'' instead of ``or'' to connect indicative words within a rule, regardless of the actual logical operator. Here, the logical relations are reflected by different operations of transformations for the disjunctive and conjunctive sub-rules. 

As to the disjunctive sub-rule, the overlapping of strong signal words is computed as:
\begin{eqnarray}
OS^{\rm d}(d,z) = \frac{SSW(d) \cap SSW(\mathcal{T}(And(r^{\rm d}(z))))}{K_2},
\end{eqnarray}
where $And(\cdot)$ is a transformation function from a logical rule to a sentence, which connects the indicative words of the rule with ``and'', i.e., $And(r^{\rm d}(z))= ``a_1~{\rm and}~a_2~{\rm and} \ldots~{\rm and}~a_S$''.

For the conjunctive sub-rule, as the involved indicative words are weaker, the matching process should be more strict. Hence, we divide the sub-rule into two parts alternately, construct the sentences separately, and take the maximum of the similarity scores:
% Formally,
\begin{eqnarray}
OS^{\rm c}(d,z)=  {\rm max} \Big( \frac{SSW(d) \cap SSW(\mathcal{T}(And(r^{\rm c1}(z)))}{K_2},\nonumber  \\
\frac{SSW(d) \cap SSW(\mathcal{T}(And(r^{\rm c2}(z)))}{K_2} \Big),
\end{eqnarray}
where $r^{\rm c1}(z) = \{b_{11}, b_{12},b_{31}, b_{32}, \ldots\}$ and $r^{\rm c2}(z) =  \{b_{21}, b_{22},b_{41}, b_{42},$ $\ldots \}$.

Finally, the similarity of word overlapping between a text $d$ and a category $z$ is defined as the sum over both sub-rules. The corresponding probability from the third perspective is then obtained through the softmax function:
\begin{equation}
OS(d,z) = OS^{\rm d}(d,z) + OS^{\rm c}(d,z),
\end{equation}
\begin{equation}
P_3(z|d) = \frac{{\rm exp}(OS(d,z))}{\sum_{z'\in Z}{\rm exp}(OS(d,z'))}.
\end{equation}

To get a final predictive probability, the three scores from different perspectives are averaged together to supplement each other:
% , especially in the weakly supervised settings.
\begin{eqnarray}
\label{eq:pseudo}
P(z|d)=(P_1(z|d)+P_2(z|d)+P_3(z|d))/3.
\end{eqnarray}
Based on this, the pseudo label of a text in the $i$-th iteration can be assigned to the category with the maximum probability:
\begin{eqnarray}
z^{(i)}(d)=\mathop{\textrm{argmax}}\limits_z(P(z|d)).
\end{eqnarray}

\begin{algorithm}[t]
\caption{RulePrompt} 
\label{alg:RulePrompt}
\begin{algorithmic}[1]
\Require
An unlabeled text corpus $D$; a set of categories $Z$ with label names; a pre-trained language model (PLM) $M$.
\Ensure
The category label $z(d)$ of each text $d\in D$.
\State Obtain initial pseudo labels $z^{(0)}(d)$ via probability distribution $P(z|d)$ for each text $d\in D$ utilizing NPPrompt with Equation~\ref{eq:Q}; 
\For{$i=1$ to $Iter$}
    \State Obtain the confidence score of each text with Equation~\ref{eq:conf};
    \State Obtain strong signal words $SSW(d)$ for each text $d\in D$ through the PLM $M$ with Equation~\ref{eq:SSW};
    \ForAll{category $z\in Z$} \Comment{Rule Mining}
        \State cluster the texts assigned to $z$ into $D^1_z, D^2_z, D^3_z$ based on their confidence scores; 
        \State Mine 1-itemsets from $D^1_z$ with Equation~\ref{eq:sup1};
        \State Mine 2-itemsets from $D^2_z$ with Equation~\ref{eq:sup2};
        \State Compose logical rule $r^{(i)}(z)$ according to Definition~\ref{def:LR};
    \EndFor
    \ForAll{text $d\in D$} \Comment{Pseudo Label Generation}
        \State Obtain new pseudo label $z^{(i)}(d)$ via probability distribution $P(z|d)$ with Equation~\ref{eq:pseudo};
    \EndFor
    \State Fine-tune the PLM $M$ with Equation~\ref{eq:finetune}; \Comment{Fine-Tuning}
\EndFor
\State \Return $z^{(Iter)}(d)$;
\end{algorithmic}
\end{algorithm}

\subsection{Self-Supervised Fine-Tuning Module}

Although prompting PLMs are strong enough to assist producing classification results in various manners, they are not specially designed for the WSTC task. Therefore, we introduce self-supervised fine-tuning into the closed loop, which uses the PLM's current prediction $P_1(d,z)$ in Equation \ref{eq:P1} to refine the PLM itself, gradually enabling it to adapt to the specific task.
Concretely, we adopt self-supervised entropy \cite{lu-etal-2022-fantastically} as the loss function to sharpen the probability distribution of category assignments generated by the PLM. That can maximize the potential of PLM and mitigate the accumulation and propagation of errors during the model training process.
Given the inaccuracy of pseudo labels, we just select a majority of texts (denoted as $D'$) with tolerable predictive probability for fine-tuning. 
Formally, the loss is defined as follows:
\begin{eqnarray}
\label{eq:finetune}
L=\sum_{d\in D'\subset D} \sum_{z\in Z} - P_1(z|d) \log P_1(z|d).
\end{eqnarray}

The fine-tuning is conducted after each main iteration updates the pseudo labels of texts,
so when the rule mining module is executed in the next iteration, new signal words derived by the fine-tuned PLM can be used.
The pseudo-codes of the overall approach is shown in Algorithm \ref{alg:RulePrompt}, and the computational complexity is analyzed in Appendix \ref{sec:complexity}.

\section{Experiments}
\label{sec:exp}

In this section, we first introduce datasets, baselines and experimental settings in the experiments. Then, overall results are presented to demonstrate the effectiveness and robustness of the proposed approach. Finally, we investigate the importance of key components by ablation study. Due to the page limit, the case study for interpretability analysis is put in Appendix \ref{sec:casestudy}, and the choices of hyperparameters will be discussed in Appendix \ref{sec:hyper}.

The experiments were performed on NVIDIA A40 GPUs, and implemented based on an open-source toolkit OpenPrompt \cite{ding2022openprompt}. 
The dataset links and codes are available on the GitHub\footnote{https://github.com/MiaomiaoLi2/RulePrompt}.

\subsection{Experimental Setup}

\subsubsection{Datasets}

We use four popular datasets from various domains for evaluation. 
The statistics of them are shown in Table \ref{tab:statistic}.
\begin{itemize}
\item \texttt{AGNews} \cite{zhang2015character} is a news article dataset from AG’s corpus.
\item \texttt{20News}\footnote{http://qwone.com/$\sim$jason/20Newsgroups/} \cite{lang1995newsweeder} is a collection of newsgroup documents.
\item \texttt{NYT} \cite{zhang2015character} contains news articles written and published by New York Times, covering abundant news topics.
\item \texttt{IMDB} \cite{maas2011learning} is for sentiment classification of movie reviews.
\end{itemize}

\begin{table}
  \caption{Dataset Statistics.}
  \label{tab:statistic}
  \Description{This table shows the statistical results of the both datasets, including XXX.}
  \begin{tabular}{ccccc}
    \toprule
   Dataset & \#Texts & \#Categories &  Classification Type &  Imbalance  \\
    \midrule
    \texttt{AGNews} & 120000 & 4 & News Topics &  1.0\\
    \texttt{20News} & 17871 & 5 & News Topics &  2.02\\
    \texttt{NYT} &  31997 & 9 & News Topics &  27.09\\
    \texttt{IMDB} &  25000 & 2 & Review Sentiment &  1.0\\
  \bottomrule
\end{tabular}
\end{table}

\subsubsection{Baselines} 

We compare our approach with the following weakly supervised  methods. 
The first two are seed-driven methods, which require at least three keywords for each category as input, and others belong to emerging PLM-based methods. 
\begin{itemize}
\item \textbf{WeSTClass} \cite{meng2018weakly} generates pseudo labels based on word embeddings and obtains the final classifier via self-training. 
\item \textbf{ConWea} \cite{mekala2020contextualized} acquires pseudo labels based on the contextualized representations of keywords, and trains a text classifier to further expand the keyword sets.
\item \textbf{LOTClass} \cite{meng2020text} utilizes the pre-trained BERT to find indicative keywords, which are directly used for category understanding and feature representation learning.
\item \textbf{XClass} \cite{wang2021x} expands indicative words for category-oriented representations, and generates pseudo labels to fine-tune a text classifier via clustering.
\item \textbf{ClassKG} \cite{zhang2021weakly}
builds a keyword graph with co-occurrence relations, and gets pseudo labels through a self-trained sub-graph annotator, used to update keywords iteratively.
\item \textbf{NPPrompt} \cite{zhao2023NPPrompt} constructs verbalizers based on initial word embeddings by PLM, and estimates the probability 
distribution over categories via weighted sum of these words.
\item \textbf{PIEClass} \cite{zhang2023PIEClass} utilizes zero-shot prompting to generate pseudo labels and improves the quality of them through two fine-tuning strategies of PLMs.
\end{itemize}

Besides, we also inspect a fully supervised method, which uses the BERT classifier with fine-tuning based on the labels in the training set. It can be regarded as an upper-bound for WSTC methods.

\subsubsection{Experimental Settings}
 
We use the standard label name of each category for each dataset as input.
As prompt-based methods are relatively robust with PLMs \cite{wang2023WSTCBenchmark}, we follow previous work \cite{zhao2023NPPrompt, wang2023PESCO} to choose RoBERTa-large \cite{liu2019roberta} as our PLM. 
The number of full iterations $Iter$ is unified to 3 across all datasets. To save space, we detail other settings and hyperparameters in Appendix \ref{sec:setting}.

As usual, we use Micro-F1 and Macro-F1 as the evaluation metrics. The results of baselines are quoted from \cite{zhang-etal-2022-prompt} 
with missing values marked as ``-''. Since NPPrompt uses more than one keyword on some datasets in its original setting, we re-run its codes provided by authors\footnote{https://github.com/XuandongZhao/NPPrompt} using only the label names for fair comparison.

\subsection{Overall Results}

\begin{table*}[h]
  \caption{Overall Results on Four Datasets by Two Metrics. The Best Scores of Weakly Supervised Methods are Marked in Bold.}
  % boldfaced and the second best score underlined.}
  % \Description{This table shows the overall results of xxx.}
  \label{tab:overall}
  \resizebox{\textwidth}{!}{
   \small
  \begin{tabular}{c|cccccccc}
    \toprule
    \multirow{2}{*}{Method} & 
    \multicolumn{2}{c}{\texttt{AGNews}} & \multicolumn{2}{c}{\texttt{20News}} & \multicolumn{2}{c}{\texttt{NYT}} & \multicolumn{2}{c}{\texttt{IMDB}} \\
    ~ & Micro-F1 & Macro-F1 & Micro-F1 & Macro-F1 & Micro-F1 &Macro-F1 & Micro-F1 &Macro-F1 \\
    \midrule
    \textbf{WeSTClass} & 0.823 &  0.821 & 0.713& 0.699 & 0.683  & 0.570  & 0.774 &- \\
    \textbf{ConWea} & 0.746 &  0.742 &  0.757 & 0.733 & 0.817  & 0.715  & -&-\\
    \textbf{LOTClass} & 0.869 & 0.868 & 0.738 & 0.725 & 0.671 & 0.436   & 0.865 & -  \\
    \textbf{XClass}  &  0.857 & 0.857 & 0.786 & 0.778 &  0.790 &  0.686 & - &- \\
    \textbf{ClassKG} &  0.881 &  0.881 & 0.811 & 0.820 & 0.721 & 0.658  & 0.888 &  0.888\\
    \textbf{NPPrompt} & 0.692 & 0.628 & 0.663 & 0.660 & 0.768 & 0.591 & 0.941 & 0.941\\
    \textbf{PIEClass (RoBERTa+RoBERTa)} &  0.895 & 0.895 &  0.755&0.760 & 0.760 & 0.694& 0.906&0.906\\
    \textbf{PIEClass (ELECTRA+ELECTRA)} &0.884 & 0.884& 0.816&0.817 & 0.832 & \textbf{0.763} &  0.931& 0.931\\
    \midrule
     \textbf{RulePrompt without Fine-Tuning} & 0.843 & 0.838  & 0.706 & 0.700 & 0.821 & 0.690 & \textbf{0.943} & \textbf{0.943} \\
     \textbf{RulePrompt} & \textbf{0.897} & \textbf{0.896} & \textbf{0.831} & \textbf{0.829} & \textbf{0.833} & 0.716 & \textbf{0.943} & \textbf{0.943}  \\
    \midrule
    \textbf{Fully Supervised} & 0.940&0.940 &  0.965 & 0.964 & 0.943 & 0.899 & 0.945 & -\\
  \bottomrule
\end{tabular}
}
\end{table*}

\begin{table*}[ht]
  \caption{Results of Ablation Study for One Iteration. The Best Scores are Marked in Bold.}
  % \Description{This table shows the overall results of xxx.}
  \label{tab:ablation}
   % \small
  \begin{tabular}{c|cccccccc}
    \toprule
    \multirow{2}{*}{Method} & 
    \multicolumn{2}{c}{\texttt{AGNews}} & \multicolumn{2}{c}{\texttt{20News}} & \multicolumn{2}{c}{\texttt{NYT}} & \multicolumn{2}{c}{\texttt{IMDB}} \\
    ~ & Micro-F1 & Macro-F1 & Micro-F1 & Macro-F1 & Micro-F1 &Macro-F1 & Micro-F1 &Macro-F1 \\
    \midrule
   \textbf{RulePrompt-1 ($- \rm Conj$) } & 0.853 & 0.850 & 0.702 & 0.694 &  0.823 & 0.709 & 0.938& 0.938\\
   \textbf{RulePrompt-1 ($- D_z$) } & 0.497 & 0.414 & 0.637 & 0.613 & 0.749 & 0.650 & 0.933 & 0.933\\   
   \midrule
    \textbf{RulePrompt-1 ($-U1$) } & 0.760 & 0.759  & 0.647 & 0.639 & 0.583 & 0.530 & 0.913 & 0.913 \\
   \textbf{RulePrompt-1 ($-U2$) } & 0.853 & 0.850 & 0.695 & 0.683 & 0.818 & 0.698 &0.937 & 0.937\\
    \textbf{RulePrompt-1 ($-U3$) } & 0.852 & 0.849 & 0.700 & 0.691 &   0.822 & 0.708 & 0.935&0.935 \\
    \midrule
    \textbf{RulePrompt-1 } & \textbf{0.854} & \textbf{0.851} & \textbf{0.705} & \textbf{0.699} &   \textbf{0.825} & \textbf{0.712}&  \textbf{0.941} & \textbf{0.941}\\
  \bottomrule
\end{tabular}
\end{table*}

The overall results of RulePrompt, its variant without fine-tuning, and baseline methods are shown in Table \ref{tab:overall}.

It is evident that our model consistently outperforms baselines for all datasets, and almost catch up with the supervised methods on \texttt{IMDB}. That certifies the role of logical rules of categories in assisting prompting PLMs to understand the topics of texts, compared with independent category-indicative words. 
In addition, the advantage over PIEClass highlights the importance of the mutual enhancement of pseudo labels and logical rules, as they are both imperfect at the starting point.
Although RulePrompt exhibits a slight gap with PIEClass on the Macro-F1 metrics in the imbalanced \texttt{NYT} dataset, which is probably caused by the amplification of categories with small samples, our approach is more stable across all datasets. 

What is more, RulePrompt significantly enhances classification accuracy on the \texttt{20News} dataset, where some categories are not completely disjoint and some label assignments are even inconsistent with general knowledge.
For example, as discussed in prior work \cite{zeng2022weakly}, 
this dataset combines ``science'' and ``encryption'' into one category, while placing ``computer'' in a separate class.
However intuitively, ``encryption'' is supposed to fall into the domain of ``science'', where ``computer'' is considered as another subset. 
That suggests our approach can effectively fuse the general knowledge embodied in the prompting PLMs and the special characteristics of the target dataset, through expressive logical rules and self-supervised fine-tuning, making it more suitable for classifying texts in challenging tasks with overlapping and counter-intuitive categories. 

Besides, the promotion over the variant without fine-tuning indicates that when there are sufficient evidences available for each category, even if unlabeled, it is still feasible to refine the PLM to accommodate the specific task and dataset, with the help of self-iterative logical knowledge of categories. However for the relatively simple and small \texttt{IMDB} dataset, adopting a fixed PLM can also achieve comparable performances.

\subsection{Ablation Study}

The ablation results for the two main modules are shown in Table \ref{tab:ablation}. In order to make the role of each component more prominent, the experiments were carried out in the first iteration (denoted as RulePrompt-1), i.e., without self-supervised fine-tuning. 

\noindent \textbf{In terms of rule mining. }
The variants include removing the conjunctive sub-rule ($- \rm Conj$), and mining rules from all texts without clustering-based set division ($- D_z$).
At first, the lack of conjunction part will lower the performance. That confirms the discrepancy among indicative words on characterizing category meanings, and the combined effect of relatively weaker words cannot be neglected.
Besides, when the rules are mined from the whole corpus, the accuracy is distinctly declined. That can be attributed to the inaccurate pseudo labels which contaminate the mining object. Therefore, the confidence scores for the predicted labels are vital to help choose appropriate texts to search for rules in an adaptive way. 

\noindent \textbf{In terms of rule-enhanced pseudo label generation. }
The variants contain the methods without either of the three units respectively.
For all cases, the full approach performs the best. That reflects different capabilities of the PLM as well as the different manners of logical rules enhancing the PLM. Since it is hard to decide which is the best one for a specific task beforehand, averaging the predictive results of them to supplement each other is a good choice, especially in the weakly supervised setting.

\section{Conclusion}

Addressing the limitations of relying solely on seed words (label names) for supervision in weakly supervised text classification task, this paper explores a kind of novel knowledge representation to characterize category meanings, which facilitates the effective integration of knowledge and unlabeled corpus. The proposed logical rules for categories can be automatically mined based on the pseudo labels of texts and iteratively self-optimized through mutual enhancement with them. Thanks to the enriched symbolic knowledge, the potential of prompting PLMs are further exploited in terms of generative capability and semantic representations, which is realized by incorporating the PLM into the rule-based iteration process. With this framework, RulePrompt exceeds the SOTA weakly supervised methods, 
and the logical rules we extract are intuitive and provide valuable guidance by disambiguating easily-confused categories.

For future work, we will strengthen the expressiveness of the category rules, such as adding the negation operator to  better avoid category confusions. Additionally, more effective iteration strategies are also worth studying, and the manner of iteratively updating pseudo labels and logical rules can be applied in other prompting PLM-based scenarios.

% \clearpage

\begin{acks}
This work is supported by CAS Project for Young Scientists in Basic Research (YSBR-040) and Strategic Priority Research Program of Chinese Academy of Sciences (XDC02060500).
\end{acks}

\bibliographystyle{ACM-Reference-Format}
\balance
\bibliography{RulePrompt}


\begin{thebibliography}{41}

%%% the Web address.

\ifx \showCODEN    \undefined \def \showCODEN     #1{\unskip}     \fi
\ifx \showDOI      \undefined \def \showDOI       #1{#1}\fi
\ifx \showISBNx    \undefined \def \showISBNx     #1{\unskip}     \fi
\ifx \showISBNxiii \undefined \def \showISBNxiii  #1{\unskip}     \fi
\ifx \showISSN     \undefined \def \showISSN      #1{\unskip}     \fi
\ifx \showLCCN     \undefined \def \showLCCN      #1{\unskip}     \fi
\ifx \shownote     \undefined \def \shownote      #1{#1}          \fi
\ifx \showarticletitle \undefined \def \showarticletitle #1{#1}   \fi
\ifx \showURL      \undefined \def \showURL       {\relax}        \fi
% invisible to TeX
\providecommand\bibfield[2]{#2}
\providecommand\bibinfo[2]{#2}
\providecommand\natexlab[1]{#1}
\providecommand\showeprint[2][]{arXiv:#2}

\bibitem[Agrawal and Srikant(1994)]%
        {agrawal1994fast}
\bibfield{author}{\bibinfo{person}{Rakesh Agrawal} {and} \bibinfo{person}{Ramakrishnan Srikant}.} \bibinfo{year}{1994}\natexlab{}.
\newblock \showarticletitle{Fast Algorithms for Mining Association Rules in Large Databases}. In \bibinfo{booktitle}{\emph{Proceedings of the 20th International Conference on Very Large Data Bases (VLDB)}}. \bibinfo{pages}{487--499}.
\newblock


\bibitem[Budd et~al\mbox{.}(2021)]%
        {budd2021survey}
\bibfield{author}{\bibinfo{person}{Samuel Budd}, \bibinfo{person}{Emma~C Robinson}, {and} \bibinfo{person}{Bernhard Kainz}.} \bibinfo{year}{2021}\natexlab{}.
\newblock \showarticletitle{A survey on active learning and human-in-the-loop deep learning for medical image analysis}.
\newblock \bibinfo{journal}{\emph{Medical Image Analysis}}  \bibinfo{volume}{71} (\bibinfo{year}{2021}), \bibinfo{pages}{102062}.
\newblock


\bibitem[Chang et~al\mbox{.}(2008)]%
        {chang2008importance}
\bibfield{author}{\bibinfo{person}{Ming-Wei Chang}, \bibinfo{person}{Lev Ratinov}, \bibinfo{person}{Dan Roth}, {and} \bibinfo{person}{Vivek Srikumar}.} \bibinfo{year}{2008}\natexlab{}.
\newblock \showarticletitle{Importance of semantic representation: dataless classification}. In \bibinfo{booktitle}{\emph{Proceedings of the AAAI Conference on Artificial Intelligence (AAAI)}}. \bibinfo{pages}{830--835}.
\newblock


\bibitem[Chen et~al\mbox{.}(2015)]%
        {chen2015dataless}
\bibfield{author}{\bibinfo{person}{Xingyuan Chen}, \bibinfo{person}{Yunqing Xia}, \bibinfo{person}{Peng Jin}, {and} \bibinfo{person}{John Carroll}.} \bibinfo{year}{2015}\natexlab{}.
\newblock \showarticletitle{Dataless text classification with descriptive LDA}. In \bibinfo{booktitle}{\emph{Proceedings of the AAAI Conference on Artificial Intelligence (AAAI)}}. \bibinfo{pages}{2224--2231}.
\newblock


\bibitem[Ding et~al\mbox{.}(2022)]%
        {ding2022openprompt}
\bibfield{author}{\bibinfo{person}{Ning Ding}, \bibinfo{person}{Shengding Hu}, \bibinfo{person}{Weilin Zhao}, \bibinfo{person}{Yulin Chen}, \bibinfo{person}{Zhiyuan Liu}, \bibinfo{person}{Haitao Zheng}, {and} \bibinfo{person}{Maosong Sun}.} \bibinfo{year}{2022}\natexlab{}.
\newblock \showarticletitle{OpenPrompt: An Open-source Framework for Prompt-learning}. In \bibinfo{booktitle}{\emph{Proceedings of the 60th Annual Meeting of the Association for Computational Linguistics: System Demonstrations}}. \bibinfo{pages}{105--113}.
\newblock


\bibitem[Fei et~al\mbox{.}(2022)]%
        {fei2022beyond}
\bibfield{author}{\bibinfo{person}{Yu Fei}, \bibinfo{person}{Zhao Meng}, \bibinfo{person}{Ping Nie}, \bibinfo{person}{Roger Wattenhofer}, {and} \bibinfo{person}{Mrinmaya Sachan}.} \bibinfo{year}{2022}\natexlab{}.
\newblock \showarticletitle{Beyond prompting: Making Pre-trained Language Models Better Zero-shot Learners by Clustering Representations}. In \bibinfo{booktitle}{\emph{Proceedings of the 2022 Conference on Empirical Methods in Natural Language Processing}}. \bibinfo{pages}{8560--8579}.
\newblock


\bibitem[Gao et~al\mbox{.}(2021)]%
        {gao-etal-2021-simcse}
\bibfield{author}{\bibinfo{person}{Tianyu Gao}, \bibinfo{person}{Xingcheng Yao}, {and} \bibinfo{person}{Danqi Chen}.} \bibinfo{year}{2021}\natexlab{}.
\newblock \showarticletitle{{S}im{CSE}: Simple Contrastive Learning of Sentence Embeddings}. In \bibinfo{booktitle}{\emph{Proceedings of the 2021 Conference on Empirical Methods in Natural Language Processing (ACL)}}. \bibinfo{pages}{6894--6910}.
\newblock


\bibitem[Han et~al\mbox{.}(2000)]%
        {han2000FreeSpan}
\bibfield{author}{\bibinfo{person}{Jiawei Han}, \bibinfo{person}{Jian Pei}, \bibinfo{person}{Behzad Mortazavi{-}Asl}, \bibinfo{person}{Qiming Chen}, \bibinfo{person}{Umeshwar Dayal}, {and} \bibinfo{person}{Meichun Hsu}.} \bibinfo{year}{2000}\natexlab{}.
\newblock \showarticletitle{{FreeSpan}: Frequent pattern-projected sequential pattern mining}. In \bibinfo{booktitle}{\emph{Proceedings of the sixth {ACM} {SIGKDD} International Conference on Knowledge Discovery and Data Mining (SIGKDD)}}, \bibfield{editor}{\bibinfo{person}{Raghu Ramakrishnan}, \bibinfo{person}{Salvatore~J. Stolfo}, \bibinfo{person}{Roberto~J. Bayardo}, {and} \bibinfo{person}{Ismail Parsa}} (Eds.). \bibinfo{publisher}{{ACM}}, \bibinfo{pages}{355--359}.
\newblock


\bibitem[Han et~al\mbox{.}(2022)]%
        {han2022ptr}
\bibfield{author}{\bibinfo{person}{Xu Han}, \bibinfo{person}{Weilin Zhao}, \bibinfo{person}{Ning Ding}, \bibinfo{person}{Zhiyuan Liu}, {and} \bibinfo{person}{Maosong Sun}.} \bibinfo{year}{2022}\natexlab{}.
\newblock \showarticletitle{{PTR}: Prompt tuning with rules for text classification}.
\newblock \bibinfo{journal}{\emph{AI Open}}  \bibinfo{volume}{3} (\bibinfo{year}{2022}), \bibinfo{pages}{182--192}.
\newblock


\bibitem[Hu et~al\mbox{.}(2022)]%
        {hu2022knowledgeable}
\bibfield{author}{\bibinfo{person}{Shengding Hu}, \bibinfo{person}{Ning Ding}, \bibinfo{person}{Huadong Wang}, \bibinfo{person}{Zhiyuan Liu}, \bibinfo{person}{Jingang Wang}, \bibinfo{person}{Juanzi Li}, \bibinfo{person}{Wei Wu}, {and} \bibinfo{person}{Maosong Sun}.} \bibinfo{year}{2022}\natexlab{}.
\newblock \showarticletitle{Knowledgeable Prompt-tuning: Incorporating Knowledge into Prompt Verbalizer for Text Classification}. In \bibinfo{booktitle}{\emph{Proceedings of the 60th Annual Meeting of the Association for Computational Linguistics (Volume 1: Long Papers) (ACL)}}. \bibinfo{pages}{2225--2240}.
\newblock


\bibitem[Hu et~al\mbox{.}(2016)]%
        {hu-etal-2016-harnessing}
\bibfield{author}{\bibinfo{person}{Zhiting Hu}, \bibinfo{person}{Xuezhe Ma}, \bibinfo{person}{Zhengzhong Liu}, \bibinfo{person}{Eduard Hovy}, {and} \bibinfo{person}{Eric Xing}.} \bibinfo{year}{2016}\natexlab{}.
\newblock \showarticletitle{Harnessing Deep Neural Networks with Logic Rules}. In \bibinfo{booktitle}{\emph{Proceedings of the 54th Annual Meeting of the Association for Computational Linguistics (Volume 1: Long Papers) (ACL)}}. \bibinfo{pages}{2410--2420}.
\newblock


\bibitem[Lang(1995)]%
        {lang1995newsweeder}
\bibfield{author}{\bibinfo{person}{Ken Lang}.} \bibinfo{year}{1995}\natexlab{}.
\newblock \showarticletitle{Newsweeder: Learning to filter netnews}.
\newblock In \bibinfo{booktitle}{\emph{Machine Learning Proceedings 1995}}. \bibinfo{publisher}{Elsevier}, \bibinfo{pages}{331--339}.
\newblock


\bibitem[Li et~al\mbox{.}(2019)]%
        {li2019filtering}
\bibfield{author}{\bibinfo{person}{Chenliang Li}, \bibinfo{person}{Shiqian Chen}, {and} \bibinfo{person}{Yan Qi}.} \bibinfo{year}{2019}\natexlab{}.
\newblock \showarticletitle{Filtering and classifying relevant short text with a few seed words}.
\newblock \bibinfo{journal}{\emph{Data and Information Management}} \bibinfo{volume}{3}, \bibinfo{number}{3} (\bibinfo{year}{2019}), \bibinfo{pages}{165--186}.
\newblock


\bibitem[Li et~al\mbox{.}(2016)]%
        {li2016effective}
\bibfield{author}{\bibinfo{person}{Chenliang Li}, \bibinfo{person}{Jian Xing}, \bibinfo{person}{Aixin Sun}, {and} \bibinfo{person}{Zongyang Ma}.} \bibinfo{year}{2016}\natexlab{}.
\newblock \showarticletitle{Effective document labeling with very few seed words: A topic model approach}. In \bibinfo{booktitle}{\emph{Proceedings of the 25th ACM International Conference on Information and Knowledge Management (CIKM)}}. \bibinfo{publisher}{{ACM}}, \bibinfo{pages}{85--94}.
\newblock


\bibitem[Li et~al\mbox{.}(2021)]%
        {li-etal-2021-weakly}
\bibfield{author}{\bibinfo{person}{Jiacheng Li}, \bibinfo{person}{Haibo Ding}, \bibinfo{person}{Jingbo Shang}, \bibinfo{person}{Julian McAuley}, {and} \bibinfo{person}{Zhe Feng}.} \bibinfo{year}{2021}\natexlab{}.
\newblock \showarticletitle{Weakly Supervised Named Entity Tagging with Learnable Logical Rules}. In \bibinfo{booktitle}{\emph{Proceedings of the 59th Annual Meeting of the Association for Computational Linguistics and the 11th International Joint Conference on Natural Language Processing (Volume 1: Long Papers) (ACL)}}. \bibinfo{address}{Online}, \bibinfo{pages}{4568--4581}.
\newblock


\bibitem[Li et~al\mbox{.}(2023)]%
        {li2023CL-WSTC}
\bibfield{author}{\bibinfo{person}{Miaomiao Li}, \bibinfo{person}{Jiaqi Zhu}, \bibinfo{person}{Xin Yang}, \bibinfo{person}{Yi Yang}, \bibinfo{person}{Qiang Gao}, {and} \bibinfo{person}{Hongan Wang}.} \bibinfo{year}{2023}\natexlab{}.
\newblock \showarticletitle{{CL-WSTC:} Continual Learning for Weakly Supervised Text Classification on the Internet}. In \bibinfo{booktitle}{\emph{Proceedings of the {ACM} Web Conference 2023 (WWW)}}. \bibinfo{publisher}{{ACM}}, \bibinfo{pages}{1489--1499}.
\newblock


\bibitem[Liao et~al\mbox{.}(2022)]%
        {liao2022zero}
\bibfield{author}{\bibinfo{person}{Chonghua Liao}, \bibinfo{person}{Yanan Zheng}, {and} \bibinfo{person}{Zhilin Yang}.} \bibinfo{year}{2022}\natexlab{}.
\newblock \showarticletitle{Zero-Label Prompt Selection}.
\newblock \bibinfo{journal}{\emph{arXiv preprint arXiv:2211.04668}} (\bibinfo{year}{2022}).
\newblock


\bibitem[Liu et~al\mbox{.}(2019)]%
        {liu2019roberta}
\bibfield{author}{\bibinfo{person}{Yinhan Liu}, \bibinfo{person}{Myle Ott}, \bibinfo{person}{Naman Goyal}, \bibinfo{person}{Jingfei Du}, \bibinfo{person}{Mandar Joshi}, \bibinfo{person}{Danqi Chen}, \bibinfo{person}{Omer Levy}, \bibinfo{person}{Mike Lewis}, \bibinfo{person}{Luke Zettlemoyer}, {and} \bibinfo{person}{Veselin Stoyanov}.} \bibinfo{year}{2019}\natexlab{}.
\newblock \showarticletitle{Roberta: A robustly optimized bert pretraining approach}.
\newblock \bibinfo{journal}{\emph{arXiv preprint arXiv:1907.11692}} (\bibinfo{year}{2019}).
\newblock


\bibitem[Loshchilov and Hutter(2018)]%
        {loshchilov2018decoupled}
\bibfield{author}{\bibinfo{person}{Ilya Loshchilov} {and} \bibinfo{person}{Frank Hutter}.} \bibinfo{year}{2018}\natexlab{}.
\newblock \showarticletitle{Decoupled Weight Decay Regularization}. In \bibinfo{booktitle}{\emph{International Conference on Learning Representations}}.
\newblock


\bibitem[Lu et~al\mbox{.}(2022)]%
        {lu-etal-2022-fantastically}
\bibfield{author}{\bibinfo{person}{Yao Lu}, \bibinfo{person}{Max Bartolo}, \bibinfo{person}{Alastair Moore}, \bibinfo{person}{Sebastian Riedel}, {and} \bibinfo{person}{Pontus Stenetorp}.} \bibinfo{year}{2022}\natexlab{}.
\newblock \showarticletitle{Fantastically Ordered Prompts and Where to Find Them: Overcoming Few-Shot Prompt Order Sensitivity}. In \bibinfo{booktitle}{\emph{Proceedings of the 60th Annual Meeting of the Association for Computational Linguistics (Volume 1: Long Papers)(ACL)}}. \bibinfo{pages}{8086--8098}.
\newblock


\bibitem[Maas et~al\mbox{.}(2011)]%
        {maas2011learning}
\bibfield{author}{\bibinfo{person}{Andrew Maas}, \bibinfo{person}{Raymond~E Daly}, \bibinfo{person}{Peter~T Pham}, \bibinfo{person}{Dan Huang}, \bibinfo{person}{Andrew~Y Ng}, {and} \bibinfo{person}{Christopher Potts}.} \bibinfo{year}{2011}\natexlab{}.
\newblock \showarticletitle{Learning word vectors for sentiment analysis}. In \bibinfo{booktitle}{\emph{Proceedings of the 49th Annual Meeting of the Association for Computational Linguistics: Human Language Technologies}}. \bibinfo{pages}{142--150}.
\newblock


\bibitem[Mekala and Shang(2020)]%
        {mekala2020contextualized}
\bibfield{author}{\bibinfo{person}{Dheeraj Mekala} {and} \bibinfo{person}{Jingbo Shang}.} \bibinfo{year}{2020}\natexlab{}.
\newblock \showarticletitle{Contextualized weak supervision for text classification}. In \bibinfo{booktitle}{\emph{Proceedings of the 58th Annual Meeting of the Association for Computational Linguistics (ACL)}}. \bibinfo{pages}{323--333}.
\newblock


\bibitem[Meng et~al\mbox{.}(2018)]%
        {meng2018weakly}
\bibfield{author}{\bibinfo{person}{Yu Meng}, \bibinfo{person}{Jiaming Shen}, \bibinfo{person}{Chao Zhang}, {and} \bibinfo{person}{Jiawei Han}.} \bibinfo{year}{2018}\natexlab{}.
\newblock \showarticletitle{Weakly-supervised neural text classification}. In \bibinfo{booktitle}{\emph{proceedings of the 27th ACM International Conference on information and knowledge management (CIKM)}}. \bibinfo{publisher}{{ACM}}, \bibinfo{pages}{983--992}.
\newblock


\bibitem[Meng et~al\mbox{.}(2020)]%
        {meng2020text}
\bibfield{author}{\bibinfo{person}{Yu Meng}, \bibinfo{person}{Yunyi Zhang}, \bibinfo{person}{Jiaxin Huang}, \bibinfo{person}{Chenyan Xiong}, \bibinfo{person}{Heng Ji}, \bibinfo{person}{Chao Zhang}, {and} \bibinfo{person}{Jiawei Han}.} \bibinfo{year}{2020}\natexlab{}.
\newblock \showarticletitle{Text classification using label names only: A language model self-training approach}. In \bibinfo{booktitle}{\emph{Proceedings of the 2020 Conference on Empirical Methods in Natural Language Processing (EMNLP)}}. \bibinfo{pages}{9006--9017}.
\newblock


\bibitem[Park et~al\mbox{.}(2023)]%
        {park2023cross}
\bibfield{author}{\bibinfo{person}{Seongmin Park}, \bibinfo{person}{Kyungho Kim}, {and} \bibinfo{person}{Jihwa Lee}.} \bibinfo{year}{2023}\natexlab{}.
\newblock \showarticletitle{Cross-task Knowledge Transfer for Extremely Weakly Supervised Text Classification}. In \bibinfo{booktitle}{\emph{Findings of the Association for Computational Linguistics: ACL 2023}}. \bibinfo{pages}{5329--5341}.
\newblock


\bibitem[Park and Lee(2022)]%
        {park2022lime}
\bibfield{author}{\bibinfo{person}{Seongmin Park} {and} \bibinfo{person}{Jihwa Lee}.} \bibinfo{year}{2022}\natexlab{}.
\newblock \showarticletitle{LIME: Weakly-Supervised Text Classification without Seeds}. In \bibinfo{booktitle}{\emph{Proceedings of the 29th International Conference on Computational Linguistics}}. \bibinfo{pages}{1083--1088}.
\newblock


\bibitem[Pei et~al\mbox{.}(2004)]%
        {pei2004PrefixSpan}
\bibfield{author}{\bibinfo{person}{Jian Pei}, \bibinfo{person}{Jiawei Han}, \bibinfo{person}{Behzad Mortazavi{-}Asl}, \bibinfo{person}{Jianyong Wang}, \bibinfo{person}{Helen Pinto}, \bibinfo{person}{Qiming Chen}, \bibinfo{person}{Umeshwar Dayal}, {and} \bibinfo{person}{Meichun Hsu}.} \bibinfo{year}{2004}\natexlab{}.
\newblock \showarticletitle{Mining Sequential Patterns by Pattern-Growth: The PrefixSpan Approach}.
\newblock \bibinfo{journal}{\emph{{IEEE} Transactions on Knowledge and Data Engineering}} \bibinfo{volume}{16}, \bibinfo{number}{11} (\bibinfo{year}{2004}), \bibinfo{pages}{1424--1440}.
\newblock


\bibitem[Scheffer et~al\mbox{.}(2001)]%
        {scheffer2001active}
\bibfield{author}{\bibinfo{person}{Tobias Scheffer}, \bibinfo{person}{Christian Decomain}, {and} \bibinfo{person}{Stefan Wrobel}.} \bibinfo{year}{2001}\natexlab{}.
\newblock \showarticletitle{Active Hidden Markov Models for Information Extraction}. In \bibinfo{booktitle}{\emph{Proceedings of the 4th International Conference on Advances in Intelligent Data Analysis}}. \bibinfo{pages}{309--318}.
\newblock


\bibitem[Song and Roth(2014)]%
        {song2014dataless}
\bibfield{author}{\bibinfo{person}{Yangqiu Song} {and} \bibinfo{person}{Dan Roth}.} \bibinfo{year}{2014}\natexlab{}.
\newblock \showarticletitle{On dataless hierarchical text classification}. In \bibinfo{booktitle}{\emph{Proceedings of the AAAI Conference on Artificial Intelligence (AAAI)}}. \bibinfo{pages}{1579--1585}.
\newblock


\bibitem[Wang et~al\mbox{.}(2023a)]%
        {wang2023PESCO}
\bibfield{author}{\bibinfo{person}{Yau{-}Shian Wang}, \bibinfo{person}{Ta{-}Chung Chi}, \bibinfo{person}{Ruohong Zhang}, {and} \bibinfo{person}{Yiming Yang}.} \bibinfo{year}{2023}\natexlab{a}.
\newblock \showarticletitle{{PESCO:} Prompt-enhanced Self Contrastive Learning for Zero-shot Text Classification}. In \bibinfo{booktitle}{\emph{Proceedings of the 61st Annual Meeting of the Association for Computational Linguistics (Volume 1: Long Papers) (ACL)}}. \bibinfo{pages}{14897--14911}.
\newblock


\bibitem[Wang et~al\mbox{.}(2021)]%
        {wang2021x}
\bibfield{author}{\bibinfo{person}{Zihan Wang}, \bibinfo{person}{Dheeraj Mekala}, {and} \bibinfo{person}{Jingbo Shang}.} \bibinfo{year}{2021}\natexlab{}.
\newblock \showarticletitle{X-Class: Text Classification with Extremely Weak Supervision}. In \bibinfo{booktitle}{\emph{Proceedings of the 2021 Conference of the North American Chapter of the Association for Computational Linguistics: Human Language Technologies}}. \bibinfo{pages}{3043--3053}.
\newblock


\bibitem[Wang et~al\mbox{.}(2023b)]%
        {wang2023WSTCBenchmark}
\bibfield{author}{\bibinfo{person}{Zihan Wang}, \bibinfo{person}{Tianle Wang}, \bibinfo{person}{Dheeraj Mekala}, {and} \bibinfo{person}{Jingbo Shang}.} \bibinfo{year}{2023}\natexlab{b}.
\newblock \showarticletitle{A Benchmark on Extremely Weakly Supervised Text Classification: Reconcile Seed Matching and Prompting Approaches}. In \bibinfo{booktitle}{\emph{Findings of the Association for Computational Linguistics: {ACL} 2023}}. \bibinfo{pages}{3944--3962}.
\newblock


\bibitem[Yang et~al\mbox{.}(2021a)]%
        {yang2021effective}
\bibfield{author}{\bibinfo{person}{Yi Yang}, \bibinfo{person}{Hongan Wang}, \bibinfo{person}{Jiaqi Zhu}, \bibinfo{person}{Wandong Shi}, \bibinfo{person}{Wenli Guo}, {and} \bibinfo{person}{Jiawen Zhang}.} \bibinfo{year}{2021}\natexlab{a}.
\newblock \showarticletitle{Effective seed-guided topic labeling for dataless hierarchical short text classification}. In \bibinfo{booktitle}{\emph{International Conference on Web Engineering (ICWE)}}. \bibinfo{pages}{271--285}.
\newblock


\bibitem[Yang et~al\mbox{.}(2021b)]%
        {yang2021dataless}
\bibfield{author}{\bibinfo{person}{Yi Yang}, \bibinfo{person}{Hongan Wang}, \bibinfo{person}{Jiaqi Zhu}, \bibinfo{person}{Yunkun Wu}, \bibinfo{person}{Kailong Jiang}, \bibinfo{person}{Wenli Guo}, {and} \bibinfo{person}{Wandong Shi}.} \bibinfo{year}{2021}\natexlab{b}.
\newblock \showarticletitle{Dataless short text classification based on biterm topic model and word embeddings}. In \bibinfo{booktitle}{\emph{Proceedings of the Twenty-Ninth International Joint Conference on Artificial Intelligence (IJCAI)}}. \bibinfo{pages}{3969--3975}.
\newblock


\bibitem[Zeng et~al\mbox{.}(2022)]%
        {zeng2022weakly}
\bibfield{author}{\bibinfo{person}{Ziqian Zeng}, \bibinfo{person}{Weimin Ni}, \bibinfo{person}{Tianqing Fang}, \bibinfo{person}{Xiang Li}, \bibinfo{person}{Xinran Zhao}, {and} \bibinfo{person}{Yangqiu Song}.} \bibinfo{year}{2022}\natexlab{}.
\newblock \showarticletitle{Weakly Supervised Text Classification using Supervision Signals from a Language Model}. In \bibinfo{booktitle}{\emph{Findings of the Association for Computational Linguistics: NAACL}}. \bibinfo{pages}{2295--2305}.
\newblock


\bibitem[Zhang et~al\mbox{.}(2021)]%
        {zhang2021weakly}
\bibfield{author}{\bibinfo{person}{Lu Zhang}, \bibinfo{person}{Jiandong Ding}, \bibinfo{person}{Yi Xu}, \bibinfo{person}{Yingyao Liu}, {and} \bibinfo{person}{Shuigeng Zhou}.} \bibinfo{year}{2021}\natexlab{}.
\newblock \showarticletitle{Weakly-supervised text classification based on keyword graph}. In \bibinfo{booktitle}{\emph{Proceedings of the 2021 Conference on Empirical Methods in Natural Language Processing (EMNLP)}}. \bibinfo{pages}{2803--2813}.
\newblock


\bibitem[Zhang et~al\mbox{.}(2022b)]%
        {zhang-etal-2022-prompt}
\bibfield{author}{\bibinfo{person}{Rongzhi Zhang}, \bibinfo{person}{Yue Yu}, \bibinfo{person}{Pranav Shetty}, \bibinfo{person}{Le Song}, {and} \bibinfo{person}{Chao Zhang}.} \bibinfo{year}{2022}\natexlab{b}.
\newblock \showarticletitle{Prompt-Based Rule Discovery and Boosting for Interactive Weakly-Supervised Learning}. In \bibinfo{booktitle}{\emph{Proceedings of the 60th Annual Meeting of the Association for Computational Linguistics (Volume 1: Long Papers) (ACL)}}. \bibinfo{pages}{745--758}.
\newblock


\bibitem[Zhang et~al\mbox{.}(2015)]%
        {zhang2015character}
\bibfield{author}{\bibinfo{person}{Xiang Zhang}, \bibinfo{person}{Junbo Zhao}, {and} \bibinfo{person}{Yann Lecun}.} \bibinfo{year}{2015}\natexlab{}.
\newblock \showarticletitle{Character-level convolutional networks for text classification}. In \bibinfo{booktitle}{\emph{Proceedings of the 28rd International Conference on Neural Information Processing Systems (NIPS)}}. \bibinfo{pages}{649--657}.
\newblock


\bibitem[Zhang et~al\mbox{.}(2022a)]%
        {zhang2022motifclass}
\bibfield{author}{\bibinfo{person}{Yu Zhang}, \bibinfo{person}{Shweta Garg}, \bibinfo{person}{Yu Meng}, \bibinfo{person}{Xiusi Chen}, {and} \bibinfo{person}{Jiawei Han}.} \bibinfo{year}{2022}\natexlab{a}.
\newblock \showarticletitle{Motifclass: Weakly supervised text classification with higher-order metadata information}. In \bibinfo{booktitle}{\emph{Proceedings of the Fifteenth ACM International Conference on Web Search and Data Mining (WSDM)}}. \bibinfo{publisher}{{ACM}}, \bibinfo{pages}{1357--1367}.
\newblock


\bibitem[Zhang et~al\mbox{.}(2023)]%
        {zhang2023PIEClass}
\bibfield{author}{\bibinfo{person}{Yunyi Zhang}, \bibinfo{person}{Minhao Jiang}, \bibinfo{person}{Yu Meng}, \bibinfo{person}{Yu Zhang}, {and} \bibinfo{person}{Jiawei Han}.} \bibinfo{year}{2023}\natexlab{}.
\newblock \showarticletitle{PIEClass: Weakly-Supervised Text Classification with Prompting and Noise-Robust Iterative Ensemble Training}. In \bibinfo{booktitle}{\emph{Proceedings of the 2023 Conference on Empirical Methods in Natural Language Processing ({EMNLP})}}. \bibinfo{publisher}{{ACM}}, \bibinfo{pages}{12655--12670}.
\newblock


\bibitem[Zhao et~al\mbox{.}(2023)]%
        {zhao2023NPPrompt}
\bibfield{author}{\bibinfo{person}{Xuandong Zhao}, \bibinfo{person}{Siqi Ouyang}, \bibinfo{person}{Zhiguo Yu}, \bibinfo{person}{Ming Wu}, {and} \bibinfo{person}{Lei Li}.} \bibinfo{year}{2023}\natexlab{}.
\newblock \showarticletitle{Pre-trained Language Models Can be Fully Zero-Shot Learners}. In \bibinfo{booktitle}{\emph{Proceedings of the 61st Annual Meeting of the Association for Computational Linguistics (Volume 1: Long Papers) (ACL)}}. \bibinfo{pages}{15590--15606}.
\newblock


\end{thebibliography}

% \clearpage

\begin{table*}[h]
  \caption{Label Names and Templates for RulePrompt.}
  \label{tab:labelnames}
  \Description{This table shows the statistical results of the both datasets, including XXX.}
  \begin{tabular}{ccc}
    \toprule
   Dataset & Label Names & Template\\
    \midrule
    \texttt{AGNews} & politics, sports, business, technology &  A [MASK] news: $d$  \\
    \texttt{20News} &computer, sports, science, politics, religion &  A [MASK] news: $d$ \\
    \texttt{NYT} & business, politics, sports, health, education, estate, arts, science, technology &  Topic: [MASK]  $d$ \\
    \texttt{IMDB} & good, bad & $d$ In summary, the film was [MASK].\\
  \bottomrule
\end{tabular}
\end{table*}

\begin{figure*}[h]
	\centering
	\subfigure[Varied Number of Iterations]{
	    \includegraphics[width=0.31\linewidth]{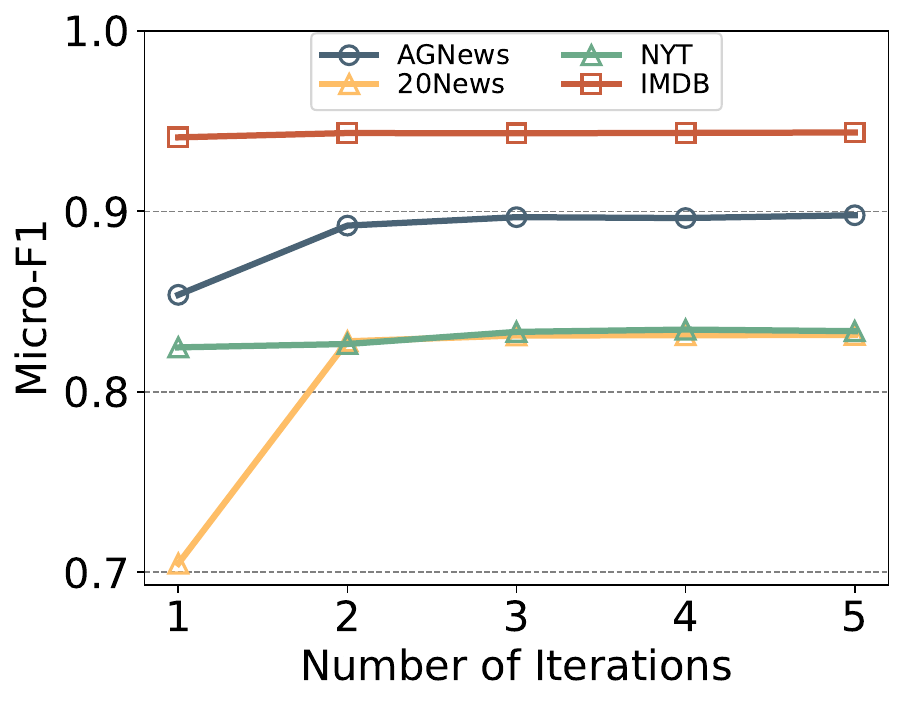}
	    \label{fig:iteration}}
	\subfigure[Varied Size of Sub-Rules]{
	    \includegraphics[width=0.31\linewidth]{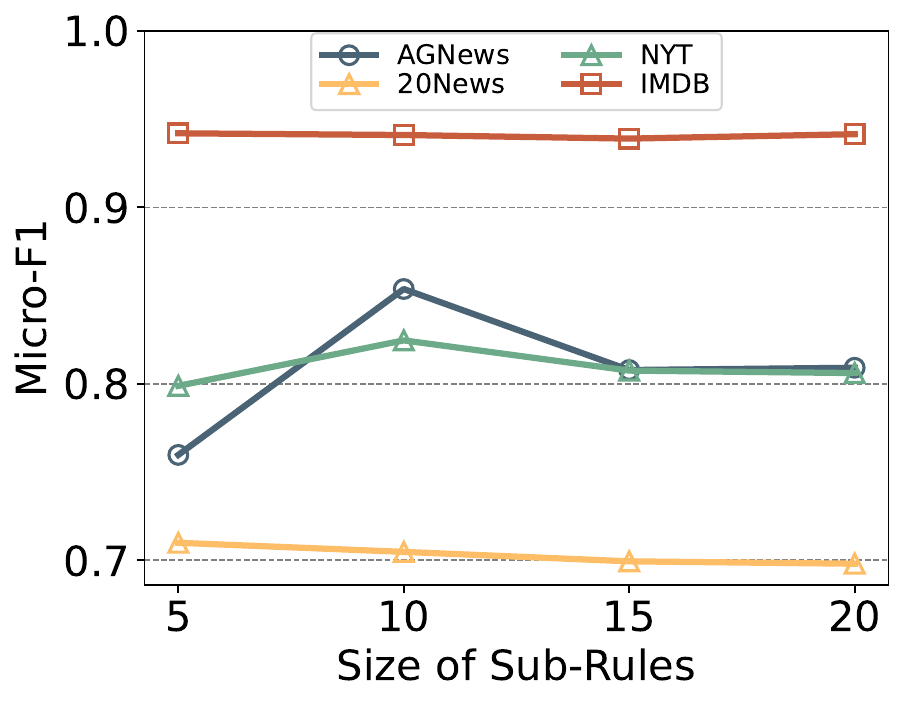}
	    \label{fig:size}}
	\subfigure[Varied Number of Strong Signal Words]{
	    \includegraphics[width=0.31\linewidth]{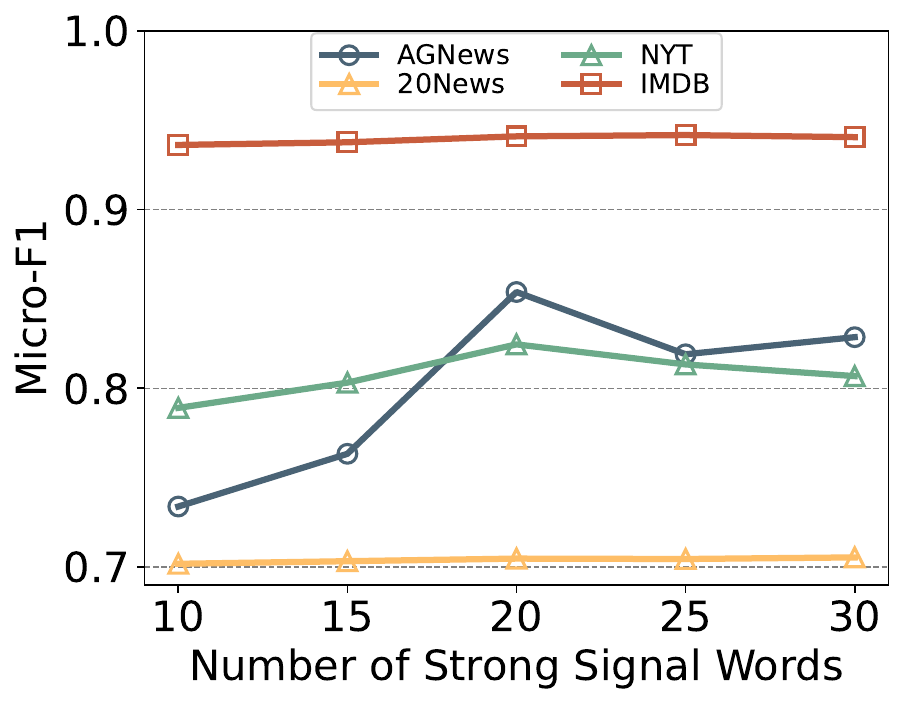}
	    \label{fig:SSW}}
	\caption{Results with Varied Hyperparameters.}
\end{figure*}

\appendix

\section{Computational Complexity}
\label{sec:complexity}

On the basis of Algorithm \ref{alg:RulePrompt}, we reckon the computational complexity of RulePrompt, assuming $N$ is the number of texts to be classified. At first, the initial pseudo labels of texts can be obtained in $O(N)$ (line 1). Then, inside each iteration, the confidence scores and the strong signal words of these texts are got in $O(N)$ (lines 3-4), and for each category, the clustering process also takes $O(N)$ (line 6). For the frequent itemset mining, since the number of items (words) is up to $O(N \cdot K_2)$, with the classical Apriori algorithm, the frequent 1-itemsets and 2-itemsets can be obtained in $O((N \cdot K_2)^2)$ (line 7) and $O((N \cdot K_2)^3)$ (line 8) respectively, and we do not need to mine long patterns. The updated pseudo labels are computed in $O(N)$ (line 12) and the fine-tuning process takes $O(N)$ with bounded text sizes. As the number threshold $K_2$ of top strong signal words, the number of categories $K$ and the iteration number $Iter$ are all constants, the overall computational complexity can be estimated as $O(N^3)$. Therefore, the proposed approach RulePrompt is scalable for larger datasets. Moreover, since the size of the sub-rules is fixed in our approach ($S=T=10$), more complex rules would not bring greater computational complexity.

\section{Detailed Experimental Settings}
\label{sec:setting}

For a fair comparison, we use the same label names of categories for each dataset as used and reported previously. Meanwhile, based on the characteristics of these datasets, we employ suitable templates according to
the previous work, and list them as well as label names in Table \ref{tab:labelnames}.

With regard to getting signal words and strong signal words of texts, we set $K_1=100$ and $K_2=20$. In the process of frequent pattern mining, we set support thresholds $h_1=h_2=0.05$ for the imbalanced \texttt{NYT}, and $0.1$ for the other three datasets. As only top 1-itemsets and 2-itemsets can enter the rule, these thresholds are insensitive and can thus be a low value. The maximum numbers of terms in the disjunctive sub-rule and conjunctions in the conjunctive sub-rule are both $S=T=10$. 
For the more complicated \texttt{20News}, we only use the word with the highest support value in the disjunctive sub-rule instead of the first half, to meet the stricter requirement for verbalizers.
In the embedding-based similarity matching unit, we choose Roberta-SimCSE \cite{gao-etal-2021-simcse} as the sentence encoder to realize the function $f(.)$.

As for the self-supervised fine-tuning process, we train 7 epochs in each iteration, except for \texttt{20News} which needs more training to understand categories, so the number of epochs is set as 30.
The learning rate is 1e-8 for \texttt{AGNews} and \texttt{20News}, while 1e-9 and 1e-10 for \texttt{NYT} and \texttt{IMDB} respectively.
The maximum sequence length is specified to 150 for \texttt{AGNews} and \texttt{NYT}, but 500 for \texttt{20News} and \texttt{IMDB}.
We use AdamW \cite{loshchilov2018decoupled} as the optimizer. 
Besides, the proportion of texts used for fine-tuning ($D'$) is set to 90\% for the imbalanced \texttt{NYT}, and 85\% for the other three datasets.

\section{Case Study}
\label{sec:casestudy}

To analyze the interpretability of logical rules derived by RulePrompt, we observe that for the ``Arts'' category in the \texttt{NYT} dataset, ``art'', ``museums'', ``galleries'', ``artwork'' and ``cultural'' are mined as 1-itemsets to constitute the disjunctive sub-rule. While, ``ballet $\land$ dancing'' and ``dancers $\land$ theater'' are identified as 2-itemsets. These paired words can indeed complement each other and form the conjunctive sub-rule. 
These rules align with common intuitions and significantly contribute to a more comprehensive representation of respective categories.
Moreover, the word ``architecture'' is found within the rules associated with two different categories: ``Estate'' and ``Arts''. It is paired with ``residential'' and ``apartments'' for the former, but ``museum'' and ``cultural'' for the latter. That exemplifies the ability of our approach to disambiguate easily-confused categories with polysemous words.

Furthermore, although the initial predictions may be incorrect, it is still beneficial for the subsequent rule mining process.
% and thus the gradual acquisition of correct results. 
Through the clustering of texts based on the pseudo labels with confidence scores, we can find appropriate strong signal words as well as their patterns to compose the rules for characterizing category meanings. During several iterations, the rules and the predictions will be optimized in the manner of mutual enhancement.
Taking the ``Business'' category in the \texttt{AGNews} dataset for example, ``x $\land$ bloomberg'' is initially mined as one of the 2-itemsets within one iteration, which is not very meaningful, but after three iterations, two new 2-itemsets ``economic $\land$ company'' and ``corporate $\land$ econom'' are mined replacing the previous one, which are more informative and consistent with the category.

\section{Hyperparameter Analysis}
\label{sec:hyper}

In this section, we pay attention to the key hyperparameters in our approach, such as the iteration number, the sub-rule size, and the number of top strong signal words, to certify the robustness of RulePrompt.

\subsection{Number of Iterations}
We vary the number of full iterations $Iter$ from 1 to 5 for all datasets, and  Figure~\ref{fig:iteration} shows the respective Micro-F1 values.
It can be seen that the performance shows a trend of first rising and then stabilizing after about three iterations. 
That is consistent across all datasets, and thus verifies our approach is insensitive on this setting as long as at least three iterations are fulfilled. 

\subsection{Size of Sub-Rules}
We make the maximum number of terms in the disjunctive sub-rule ($S$) and the maximum number of conjunctions in the conjunctive sub-rule ($T$) equal to each other, and vary them together between 5 and 20 with the step size 5. The results in Figure \ref{fig:size} show a trend of first rising and then declining, with an optimal value of 10, which is nearly consistent for all datasets. 

An exception appears for the complicated dataset $\texttt{20News}$, where the accuracy always decreases mildly along with the increase of sub-rule sizes.
This is mainly because some categories in $\texttt{20News}$ are largely overlapping, and the stricter rules are thereby required to distinguish them. 

\subsection{Number of Strong Signal Words}
We change the number threshold $K_2$ of top strong signal words between 10 and 30 with the step size 5. The performances are shown in Figure \ref{fig:SSW}. 
Again, an almost optimal value 20 is reached for all datasets with a similar trend as the size of sub-rules. 

Note that another threshold $K_1$ is used to determine a larger candidate signal word set, so does not directly affect the results and behaves insensitive obviously.

\end{document}